\documentclass{article}
\usepackage{ijcai26}

\usepackage{times}
\usepackage{soul}
\usepackage{url}
\usepackage{xcolor}
\usepackage{colortbl}
\usepackage[hidelinks]{hyperref}
\usepackage[utf8]{inputenc}
\usepackage[small]{caption}
\usepackage{graphicx}
\usepackage{amsmath}
\usepackage{amsthm}
\usepackage{booktabs}
\usepackage{algorithm}
\usepackage{algorithmicx}
\usepackage{algpseudocode}
\usepackage{booktabs}
\usepackage{adjustbox}
\usepackage{multirow}
\usepackage{amsmath, amsfonts, amssymb}
\usepackage[switch]{lineno}
\usepackage{array}
\usepackage{threeparttable}
\newcolumntype{P}[1]{>{\centering\arraybackslash}p{#1}}
\usepackage[caption=false, font=footnotesize]{subfig}

\title{BehaviorGuard: Online Backdoor Defense for Deep Reinforcement Learning\footnote{Accepted in IJCAI 2026}}

\author{
Yinbo Yu$^1$\and
Xueyu Yin$^2$\and
Jiadai Wang$^2$\and
Chunwei Tian$^3$\and
Sai Xu$^4$\and
Qi Zhu$^1$\and
Daoqiang Zhang$^1$\\
\affiliations
$^1$College of Artificial Intelligence, Nanjing University of Aeronautics and Astronautics\\
$^2$School of Cybersecurity, Northwestern Polytechnical University\\
$^3$School of Computer Science and Technology, Harbin Institute of Technology\\
$^4$Department of Electronic and Electrical Engineering, University College London\\
\emails
yinboyu@nuaa.edu.cn,
xueyuy@mail.nwpu.edu.cn,
sai.xu@ucl.ac.uk
}

\begin{document}

\maketitle

\begin{abstract}
Backdoor attacks pose a serious threat to deep reinforcement learning (DRL). Current defenses typically rely on reward anomalies to reverse-engineer triggers and model finetuning to remove backdoors. However, complex trigger patterns undermine their robustness, and fine-tuning entails high costs, limiting practical utility. Therefore, we shift defense concerns to trigger-agnostic backdoor output behaviors and propose BehaviorGuard\footnote{ \url{https://github.com/c0d818/BehaviorGuard}}, an online behavior-based backdoor detection and mitigation framework for DRL. Specifically, we find that regardless of attacks, backdoored policies induce consistent shifts in action distributions to ensure reliable activation, leaving detectable traces in high-quantile regions and distribution tails, even in the absence of triggers. Based on this, we design a novel metric that captures behavioral drift in action distributions to identify and suppress backdoor actions at runtime. To our knowledge, this is the first online backdoor defense that counters attacks both in single- and multi-agent DRL. Evaluated across diverse benchmarks with different backdoor attacks, BehaviorGuard consistently surpasses prior methods in both efficacy and efficiency.


\end{abstract}
\section{Introduction}
Deep reinforcement learning (DRL) has shown strong performance in complex sequential decision-making tasks and is starting to be used in safety-critical domains such as games \cite{vinyals2019grandmaster}, autonomous driving \cite{wang2021stop}, and robotics \cite{tang2025deep}. As DRL moves into such settings, security and safety become central concerns \cite{gu2024review}. Recent studies \cite{kiourti2019trojdrl,wang2021backdoorl} have revealed that DRL policies are highly vulnerable to backdoor attacks, where an adversary implants a hidden backdoor during policy training that causes the agent to behave maliciously once the backdoor is activated, while maintaining benign performance otherwise. Compared to backdoors in supervised learning \cite{gu2017badnets,liu2018trojaning}, backdoors in DRL are particularly dangerous due to the sequential nature of decision making and complex environment dynamics, which can amplify small behavior shifts into catastrophic outcomes.

This threat is further amplified in multi-agent reinforcement learning (MARL) settings. In competitive environments, an adversary may trigger a victim’s backdoor purely through its own actions or interaction patterns \cite{wang2021backdoorl}, whereas in cooperative environments, poisoning a single agent can compromise the entire team through coordination failure and cascading effects \cite{chen2022marnet}. The inherent non-stationarity and inter-agent coupling in MARL significantly increase the difficulty of detecting and mitigating backdoors, making defenses developed for single-agent settings unreliable when transferred across environments.

Existing defenses against DRL backdoors primarily rely on reward anomalies, trigger reconstruction, or prior knowledge of trigger patterns \cite{guo2022policycleanse,chen2023bird,yuan2024shine}. 
Although these methods show high defensive capabilities, they still face several challenges: (1) reward signals may appear normal when the backdoor is active \cite{rathbun2025adversarial}, resulting in a high detection false-negative rate;
(2) pattern-dependent methods fail against triggers based on sequential behaviors with spatial or temporal dependence \cite{yu2023spatiotemporal}; (3) most methods lack a unified treatment that generalizes across both single- and multi-agent scenarios; and (4) many methods require expensive policy retraining to mitigate backdoors.

To address these challenges, in this work, we shift our defense focus from recognizing diverse and attack-specific trigger patterns targeted by prior work to identifying the unified backdoor output manifestation. Specifically, we analyze the output behavioral differences between clean and backdoor DRL policies. Our key observation is that, regardless of trigger patterns or environments, backdoored policies inevitably induce subtle but persistent behavioral drift in the action distribution. Even under trigger-free inputs, the policy must bias its action probabilities to ensure successful activation, leaving detectable signatures in high quantiles and tails of the distribution. This insight enables trigger-agnostic backdoor defense without relying on reward signals or trigger priors.

Building on this finding, we propose \textbf{BehaviorGuard}, a unified framework that jointly detects and mitigates backdoors online for different DRL scenarios. We first design a backdoor detection method based on the signature of behavioral drift, which quantifies action-distribution deviations using the Behavioral Drift Score (BDS) and density-based statistics. Second, we propose a lightweight, retraining-free, drift-constrained backdoor mitigation strategy to mitigate backdoor actions online at runtime. Once backdoor outputs are detected, BehaviorGuard suppresses backdoor behaviors by projecting the abnormal action distribution back to a low-drift region, achieving practical protection without retraining. To the best of our knowledge, this is the first work that can detect and mitigate backdoor attacks against single- and multi-agent DRL at runtime. 

Our contributions are summarized as follows:
\begin{itemize}
    \item We identify behavioral drift as an attack- and environment-agnostic Trojan signature.
    \item We propose a unified backdoor detection framework for single- and multi-agent environments.
    \item We introduce a lightweight drift-constrained mitigation method that suppresses backdoor behaviors at inference time without retraining, enabling online defense.
    \item Extensive experiments across single-, cooperative, and competitive multi-agent benchmarks demonstrate that BehaviorGuard achieves powerful detection performance while effectively mitigating backdoor impact with minimal degradation on clean performance.
\end{itemize}

\section{Background and Related Work}

\subsection{Deep Reinforcement Learning}
Single-agent RL (SARL) is commonly modeled as an MDP $\mathcal{M}=(\mathcal{S},\mathcal{A},P,r,\gamma)$ , where $S$ and $A$ denote the state and action spaces, $P(s^\prime|s,a)$ is the transition kernel, $r(s,a)$ is the immediate reward, and $\gamma\in(0,1)$ is the discount factor. At time $t$, the agent observes a state (or observation) $x_t$ and samples an action $a_t\sim\pi_\theta(\cdot|x_t)$; the environment then transitions to $s_{t+1}$ and returns reward $r_t$. The policy $\pi_\theta$ is typically parameterized by a deep neural network (DNN), and is trained to maximize the expected discounted return $\mathbb{E}_{\pi_\theta}[\sum^{\infty}_{t=0}\gamma^tr(s_t,a_t)]$.

Beyond the single-agent setting, we also consider multi-agent RL (MARL) \cite{busoniu2008comprehensive}. In competitive MARL (com-MARL), agents’ rewards are competitive or conflicting, and the environment evolves under the joint action $a_t=(a^1_t,\dots,a^n_t)$ \cite{lowe2017multi}. In cooperative MARL (coo-MARL), agents share a team objective (e.g., a shared team reward), where coordination among policies is essential \cite{rashid2020monotonic,foerster2018counterfactual}. 

\subsection{Backdoor Attack Formalization in DRL}
\begin{figure}[t!]
\centering
\begin{minipage}[c]{1\linewidth}
    \centering
    \subfloat[Instant Trigger]{\includegraphics[width=0.36\linewidth]{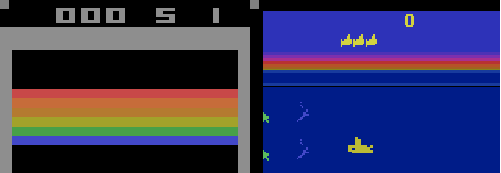}}
    \hspace{1mm}
    \subfloat[Sequential Trigger]{\includegraphics[width=0.57\linewidth]{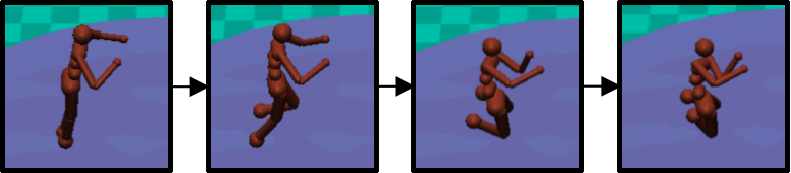}}
\end{minipage}
\vspace{-3mm}
\caption{(a) A 3×3 pixel patch in the upper left corner; (b) A sequence of actions over 4 consecutive timesteps.}
\label{fig:trigger}
\vspace{-5mm}
\end{figure}

Backdoor attacks aim to implant a hidden malicious behavior into a policy network such that it behaves similarly to a clean policy under normal inputs, while reliably switching to an attacker-specified behavior when a trigger is present. Formally, let $\delta$ denote the trigger and $f_\delta(\cdot)$ be the trigger injection function that applies the trigger pattern to an input: $x^\delta_t=f_\delta(x_t)$.
As shown in Fig. \ref{fig:trigger}, triggers can be classified into two types: \textbf{instant} and \textbf{sequential} triggers. The former can be instantiated as single-step state perturbations (e.g., visual patches \cite{kiourti2019trojdrl} or specific states \cite{wang2021stop}). The latter can be temporal state patterns \cite{yu2022temporal} or interaction-based conditions among agents \cite{wang2021backdoorl,fang2025blast} spanning multiple steps. The attacker obtains a backdoored policy $\pi^{bd}$ via poisoning the training process such that $\pi^{bd}$ mimics a clean policy $\pi^{cl}$ on trigger-free inputs but shifts toward a target behavior $\pi^{tar}$ on triggered inputs:
\vspace{-1mm}
\begin{equation}
\pi^{bd}(\cdot|x)\approx\pi^{cl}(\cdot|x),\quad \pi^{bd}(\cdot|f_\delta(x))\approx\pi^{tar}(\cdot|f_\delta(x)).
\label{pi}
\end{equation}

In SARL, this switch often manifests as a sharp return drop or persistent target actions; in com-MARL, the attack may reduce the victim’s win rate or induce systematic failures in key adversarial states; in coo-MARL, deviations of one agent can further disrupt coordination and degrade team performance.

\subsection{Backdoor Attacks in DRL}

\textbf{Backdoor Attacks in SARL.} TrojDRL \cite{kiourti2019trojdrl} is the first to show backdoor attacks on deep RL agents, demonstrating that a small amount of poisoned data and reward tweaks could implant malicious behavior. Yu et al. \cite{yu2022temporal} introduce a new DRL backdoor whose trigger is a set of sequential observations, making it stealthy and persistent. BadRL \cite{cui2024badrl} uses highly sparse poisoned samples during training and testing, maintaining a high attack success rate while reducing cost. BAFFLE \cite{gong2024baffle} implants a backdoor by poisoning offline RL datasets.

\textbf{Backdoor Attacks in com-MARL.} BACKDOORL \cite{wang2021backdoorl} uses a sequence of actions by the opponent as a trigger, so that the adversary can activate the victim agent's backdoor through its own actions. SCAB \cite{liu2025fox} injects backdoors via pre-trained agents interacting with legitimate actions without the need for observation perturbations or reward modification.

\textbf{Backdoor Attacks in coo-MARL.} MARNet \cite{chen2022marnet} successfully implements backdoor attacks on coo-MARL systems by designing trigger, action poisoning, and reward hacking modules, causing the attacker's agent to behave abnormally in a malicious environment. One4all\cite{zheng2023one4all} presents two poisoning attack techniques (SNPA and TAPA), which covertly poison the coo-MARL system by manipulating a single agent's observations or reward functions. BLAST \cite{fang2025blast} introduces a unilateral filter that enables a single backdoor agent to quickly affect other clean agents, thereby enabling a bakdoor leverage attack on the entire multi-agent system. 

\subsection{Backdoor Defenses in DRL}

Acharya et al. \cite{acharya2023universal} detect Trojan agents by analyzing discrepancies in how the agent's policy evaluates state observations. Vyas et al. \cite{vyas2024mitigating} present a runtime detection method based on neural activation patterns, effectively detecting subtly hidden backdoor triggers. These methods can detect backdoor attacks in DRL, but cannot mitigate those backdoors. PolicyCleanse \cite{guo2022policycleanse} detects Trojan agents in com-MARL according to abnormal accumulated rewards and mitigates the backdoor via machine unlearning. BIRD \cite{chen2023bird} and SHINE \cite{yuan2024shine} leverage trigger restoration and policy finetuning to detect and remove backdoors in DRL policies without requiring attack specifications. These methods all rely on reward anomalies and expensive policy retraining to detect and mitigate backdoors, which results in potential missed detection risks and high defense overhead. Bharti et al. \cite{bharti2022provable} proposed a provable, retraining-free backdoor defense mechanism that projects observations onto a safe subspace to suppress backdoor activation. However, it only supports the single-agent scenario. 

Different from current methods, our BehaviorGuard is based on a novel behavioral drift metric without trigger priors, reward signals, or policy retraining, and can be deployed to defend against backdoors online. Moreover, it is effective against different trigger patterns, instant or sequential behaviors, environments (SARL, com-MARL or coo-MARL) and retains defense power even when rewards appear normal.  
\section{Method}
\subsection{Threat Model}
\textbf{Attacker's Capabilities and Goal}.
The attacker has complete control over the training process, data, and reward functions, and can inject any backdoors with instant or sequential triggers into policies. These policies can strive to achieve their benign objectives, but once the backdoor is activated, they will immediately perform backdoor objectives.

\textbf{Defender's Capabilities and Goal}.
The defender can only download or outsource the final policy, which may potentially contain a backdoor. We assume the defender does not have the resources to retrain policies. This enables higher practicality. The defender aims to detect whether the policy contains a backdoor and, if so, mitigate its impact.



\subsection{Key Intuition}
\begin{figure}[t]
\centering
\begin{minipage}[c]{1\linewidth}
    \centering
    \subfloat[SARL: Breakout]{\includegraphics[width=0.46\linewidth]{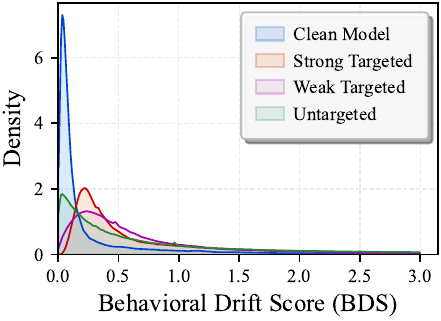}}
    \hspace{2mm}
    \subfloat[com-MARL: sumo-humans]{\includegraphics[width=0.46\linewidth]{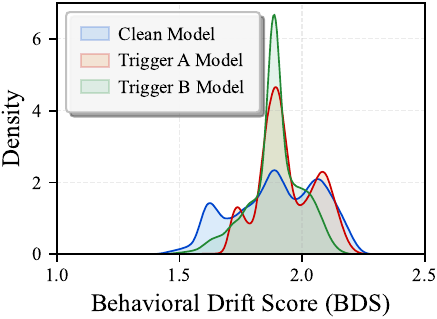}}
\end{minipage}
\vspace{-3mm}
\caption{Behavioral drift comparison of clean and backdoor policies in single- and multi-agent environments.}
\label{fig:drift_compare}
\vspace{-4mm}
\end{figure}

One key observation across different backdoor attacks (e.g., TrojDRL\cite{kiourti2019trojdrl} and BACKDOORL\cite{wang2021backdoorl}) is that their policies systematically reshape the action distribution, leaving detectable \textit{behavioral drifts}. As shown in Fig. \ref{fig:drift_compare}, which compares the BDS distribution of clean policies and those injected with 3 types of TrojDRL attacks in SARL and 2 types of BACKDOORL attacks in com-MARL, these drifts manifest in distinct patterns. In SARL, the clean BDS distribution is highly concentrated, owing to the stability of the single-agent environment. Backdoor attacks undermine this consistency, resulting in dispersed distributions with heavier tails. In contrast, in the non-stationary environment of com-MARL, a clean policy naturally exhibits a wider BDS distribution due to adversarial interactions. Intriguingly, BACKDOORL counteracts this by enforcing a fixed, cooperative strategy to achieve its attack goal, leading to an abnormally concentrated BDS distribution. In summary, despite variations in its directional pattern across environments, \textit{the behavioral drift induced by the inherent conflict between primary and backdoor tasks serves as a robust, universal clue for backdoor defense}.

\subsection{Behavioral Drift Detection for Backdoor Detection}
Based on the above observation, in BehaviorGuard, we design a behavioral drift detection (BDD) method based on BDS to detect backdoors against both SARL and MARL. However, their underlying mechanisms and observable characteristics differ substantially. Hence, to maintain conceptual clarity and analytical focus, here, we mainly describe our designs of BDD for SARL and leave its extensions for coo- and com-MARL in Supplement A.


In the single-agent scenario, once the backdoor is activated, it will induce systematic shifts in the policy’s action distributions under attack conditions, while the policy remains indistinguishable from a clean model on benign inputs. BDD aims to identify these attacks by quantifying deviations between the policy’s runtime behavior and its expected benign statistics. Given a trajectory of length $T_s$, BDS at each timestep $t$ can be calculated as:
\begin{equation}
\text{BDS}_{t}=\frac{1}{2}\sum_{k=1}^{K}\left(\frac{p_{t,k}-\mu_{k}}{\sigma_{k}+\epsilon}\right)^2,
\label{BDS}
\end{equation}
where $p_{t,k}$ denotes the predicted probability of the $k$-th action at timestep $t$, and $(\mu_k,\sigma_k)$ are the corresponding mean and standard deviation estimated from clean executions. The constant $\epsilon$ is used for
numerical stability.


To characterize the distribution of behavioral drift over the entire execution, we collect the sequence $\{\text{BDS}_t\}_{t=1}^{T_s}$ and estimate its probability density function using KDE \cite{scott2015multivariate}:
\begin{equation}
\hat{f}(\text{BDS}_{t})=\frac{1}{Nh}\sum_{i=1}^{N}K\left(\frac{\text{BDS}_{t}-\text{BDS}_{i}}{h}\right),
\label{PDF}
\end{equation}
where $K(\cdot)$ denotes a Gaussian kernel, $h$ is the bandwidth parameter, and
$\{\text{BDS}_i\}_{i=1}^{N}$ are the observed drift samples used for density estimation. Based on the estimated density, we define a \emph{tail statistic} that measures the proportion of timesteps exhibiting rare, low-density deviations:
\begin{equation}
\text{tail}=\frac{1}{T_s}\sum_{t=1}^{T_s}
\mathbf{1}\left[\hat{f}(\text{BDS}_t)<\tau_{\text{tail}}\right],
\label{tail}
\end{equation}
where $\tau_{\text{tail}}$ is defined as the density value corresponding to the low-density tail of the BDS distribution. It is calibrated by fitting a KDE to the clean baseline $\text{BDS}_t$ values and setting $\tau_{\text{tail}}$ to the 5th percentile of the estimated densities $\hat f(\text{BDS}_t)$.


We further aggregate the drift magnitude over the trajectory by computing the mean drift score $\overline{\text{BDS}}=\frac{1}{T_s}\sum_{t=1}^{T_s}\text{BDS}_t$.
Both $\overline{\text{BDS}}$ and $\text{tail}$ are normalized using Z-scores:
\begin{equation}
z_{\text{BDS}}=\frac{\overline{\text{BDS}}-\mu_{\text{BDS}}}
{\max(\sigma_{\text{BDS}},\epsilon)},\quad
z_{\text{tail}}=\frac{\text{tail}-\mu_{\text{tail}}}
{\max(\sigma_{\text{tail}},\epsilon)}.
\label{zBDS}
\end{equation}
Finally, they are combined into a composite detection score:
\begin{equation}
\widehat{\text{BDS}}=\frac{1}{\sqrt{2}}(z_{\text{BDS}}+z_{\text{tail}}),
\label{z}
\end{equation}
where we use $\frac{1}{\sqrt{2}}$ to keep $\widehat{\text{BDS}}$'s range being consistent with $\text{BDS}$. A policy execution is flagged as potentially backdoored when $\widehat{\text{BDS}}$ exceeds a predefined threshold $\tau$. We set $\tau$ as the 95th percentile of the composite score distribution computed from clean baseline executions, and fix it for all subsequent backdoor detection. The complete detection procedure is summarized in Algorithm~\ref{alg:sarl-bdd}. 

\begin{algorithm}[t]
\caption{Behavioral Drift Detection of BehaviorGuard}
\label{alg:sarl-bdd}
\small
\begin{algorithmic}[1]
\Require Action probabilities $\{p_{t,k}\}_{t=1}^{T_s}$; benign stats $(\mu_k,\sigma_k)$; KDE $(K,h)$; density threshold $\tau_{\text{tail}}$; weights $(w_1,w_2)$; decision threshold $\tau$
\Ensure Detection result
\State Initialize $\mathcal{B}\gets [\,]$ 
\For{$t=1$ to $T_s$}
    \State $\text{BDS}_t \gets \frac{1}{2}\sum_{k=1}^{K}\left(\frac{p_{t,k}-\mu_{k}}{\sigma_{k}+\epsilon}\right)^2$ 
    \State Append $\text{BDS}_t$ to $\mathcal{B}$
\EndFor
\For{$t=1$ to $T_s$}
    \State $\hat f(\mathcal{B}[t]) \gets \frac{1}{Nh}\sum_{i=1}^{N}K\!\left(\frac{\mathcal{B}[t]-\mathcal{B}[i]}{h}\right)$ 
\EndFor
\State $\text{tail} \gets \frac{1}{T_s}\sum_{t=1}^{T_s}\mathbf{1}\!\left[\hat f(\mathcal{B}[t])<\tau_{\text{tail}}\right]$ 
\State $\overline{\text{BDS}} \gets \frac{1}{T_s}\sum_{t=1}^{T_s}\mathcal{B}[t]$
\State $z_{\text{BDS}} \gets \frac{\overline{\text{BDS}}-\mu_{\text{BDS}}}{\max(\sigma_{\text{BDS}},\epsilon)}$, \;
       $z_{\text{tail}} \gets \frac{\text{tail}-\mu_{\text{tail}}}{\max(\sigma_{\text{tail}},\epsilon)}$ 
\State $\text{score} \gets \frac{1}{\sqrt{2}} (z_{\text{BDS}} + z_{\text{tail}})$
\If{$\text{score}>\tau$}
    \State \Return ``Potential Backdoor Detected''
\Else
    \State \Return ``No Backdoor Detected''
\EndIf
\end{algorithmic}
\end{algorithm}
\setlength{\textfloatsep}{10pt}

The above detection formulation is designed for single-agent policies, and extending it to multi-agent settings requires additional considerations. In \emph{competitive} multi-agent environments, backdoor behaviors often emerge through interaction-dependent deviations, where an agent’s abnormal actions are amplified by the opponent’s responses. In contrast, \emph{cooperative} multi-agent scenarios exhibit correlated behavior shifts across agents, in which backdoor activation may manifest as synchronized or role-specific deviations. These characteristics render direct application of single-agent BDD insufficient and require dedicated designs that account for inter-agent dependency and joint behavioral dynamics. Those details of the extended BDD are described in Supplement A. 

\subsection{Behavioral-Drift–Constrained Mitigation}
In the spirit of runtime shielding for RL \cite{alshiekh2018safe}, we further design an online backdoor mitigation method in BehaviorGuard based on behavioral drift, which constrains policy execution by leveraging the output of BDD. This method is summarized in Algorithm \ref{alg:dg-action-guard}. Without requiring trigger priors or policy retraining, our method intervenes online to suppress abnormal actions induced by persistent deviations from normal policy execution.

\begin{algorithm}[t]
\caption{Drift-Constrained Backdoor Action Mitigation}
\label{alg:dg-action-guard}
\small
\begin{algorithmic}[1]
\Require 
Victim policy $\pi_\theta$; drift detector producing score $\text{BDS}_t$; threshold $\tau_D$; \\
reference action sequence $\mathcal{A}^{\mathrm{ref}}=\{a_t^{\mathrm{ref}}\}_{t=1}^{T}$; 
window length $L$; guard probability $p$
\Ensure Executed action $\tilde a_t$
\State Initialize drift indicator buffer $\{z_i\}$ with length $L$ to zeros
\For{$t = 1,2,\dots$}
    \State Observe current state $s_t$
    \State Compute raw action $a_t \gets \arg\max_a \pi_\theta(a \mid s_t)$
    \State Obtain reference action $a_t^{\mathrm{ref}} \gets \mathcal{A}^{\mathrm{ref}}[t \bmod T]$
    \State Obtain drift score $\text{BDS}_t$ from detector
    \State $z_t \gets \mathbb{I}(\text{BDS}_t > \tau_D)$
    \State Update buffer and compute $c_t \gets \sum_{i=\max(1,t-L+1)}^{t} z_i$
    \State $g_t \gets \mathbb{I}(c_t \ge L)$
    \If{$g_t = 1$ \textbf{and} $\mathrm{Unif}(0,1) < p$}
        \State $\tilde a_t \gets a_t^{\mathrm{ref}}$
    \Else
        \State $\tilde a_t \gets a_t$
    \EndIf
    \State Execute action $\tilde a_t$ and observe next state
\EndFor
\end{algorithmic}
\end{algorithm}
\setlength{\textfloatsep}{10pt}

In the single-agent environment, we assume that typical action distribution characteristics under benign conditions can be approximated from limited offline interactions or historical execution logs, and use them as a normal behavior prior that remains stable under similar environment configurations. Formally, the prior is represented as a time-indexed set of reference actions:
\begin{equation}
\mathcal{A}^{\mathrm{ref}}=\{a_t^{\mathrm{ref}}\}_{t=1}^{T}, 
\label{eq:A_ref}
\end{equation}
where $a^{ref}_t$ denotes the most representative action at timestep $t$ under benign conditions (e.g., the mode of the action distribution). During policy inference, the reference action is retrieved via modular indexing
\begin{equation}
a^\mathrm{ref}_t=\mathcal{A}^\mathrm{ref}[t\;\text{mod}\;T], \label{a^ref}
\end{equation}
which provides a lightweight behavioral reference without requiring exact state alignment.

At each timestep, BDD outputs a drift score $\text{BDS}_t$. When it exceeds a predefined threshold $\tau_D$, the output is considered to deviate from normal behavior. $\tau_D$ is calibrated offline as the 95th percentile of the BDS distribution obtained from clean baseline executions, thus filtering out innocuous noise.
To reduce false triggers caused by stochasticity or transient noise, we smooth the detection results over time by defining a drift indicator:
\begin{equation}
z_t = \mathbb{I}(\text{BDS}_t > \tau_D), \label{z_t}
\end{equation}
where $\mathbb{I}(\cdot)$ denotes the indicator function. Drift occurrences are then accumulated within a temporal window of length $L$:
\begin{equation}
c_t = \sum_{i=\max(1,t-L+1)}^{t} z_i, 
\label{eq:c_t}
\end{equation}
where $L$ controls how many consecutive drift detections are required before mitigation is triggered. When $c_t \ge L$, the policy is regarded as entering a persistent abnormal behavior regime, and a mitigation gate $g_t=\mathbb{I}(c_t \ge L)$ is activated.

At timestep $t$, the policy produces a raw action:
\begin{equation}
a_t = \arg\max_a \, \pi_\theta(a \mid s_t). 
\label{eq:a_t}
\end{equation}
If mitigation is not triggered ($g_t=0$), the raw action is executed. Otherwise, when persistent behavioral drift is detected ($g_t=1$), the action is probabilistically corrected with probability $p$ by projecting it back to the reference action:
\begin{equation}
\tilde a_t =
\begin{cases}
a_t^{\mathrm{ref}}, & \text{if } g_t = 1 \text{ and } u < p,\\
a_t, & \text{otherwise},
\end{cases}
\quad u \sim \mathrm{Unif}(0,1).
\label{eq:tilde_a_t}
\end{equation}
$p$ is a mitigation probability. This probabilistic correction avoids deterministic intervention at all timesteps and allows the mitigation strength to be controlled via parameters $p$ and $L$, thereby suppressing abnormal behavior while preserving benign policy performance.

Overall, BehaviorGuard forms a unified closed loop between detection and mitigation: BDD determines whether the policy has entered an abnormal regime, while action constraints intervene at the execution level only when persistent drift is confirmed. This design enables effective and controllable backdoor suppression with a low computational overhead, without policy retraining or trigger knowledge. Extensions to multi-agent environments based on the same principle are provided in the Supplement A.

\section{Evaluation}


\subsection{Experiment Setup}
\textbf{Evaluation Environments.}
We evaluate the performance of BehaviorGuard against backdoor attacks in discrete-action settings (SARL and coo-MARL) and a continuous-action setting (com-MARL).

\textit{SARL.}
We follow TrojDRL \cite{kiourti2019trojdrl} and consider three attack variants: Strong-Targeted, Weak-Targeted, and Untargeted. Experiments are conducted on four Atari games from OpenAI Gym \cite{brockman2016openai} and ALE \cite{bellemare2013arcade}: Breakout, Pong, Seaquest, and SpaceInvaders.
These tasks exhibit diverse dynamics and action-selection patterns, allowing us to evaluate detection behavior across heterogeneous state distributions.

\emph{com-MARL.}
We adopt BACKDOORL \cite{wang2021backdoorl}, which targets adversarial interaction settings, and evaluate on three tasks from the competitive benchmark suite \cite{bansal2017emergent}: Sumo-Humans, Run-To-Goal, and You-Shall-Not-Pass.
The strategic coupling and non-stationarity induced by opponent behaviors make this setting particularly challenging for trigger-agnostic detection.

\emph{coo-MARL.}
We evaluate against the MARNet attack \cite{chen2022marnet} on two representative coo-MARL algorithms (QMIX \cite{QMIX} and COMA \cite{foerster2018counterfactual}), which uses specific textures as its trigger. 


We set the mitigation probability $p=0.5$ and the window length $L=5$ to default across all environments. For each task, we evaluate multiple clean and backdoored policies to account for training variability.

\textbf{Metrics.} We assess the proposed approach using key metrics. We use the average episode reward in SARL tasks and the winning rate in MARL tasks to compare different policies. We use the False Positive Rate (FPR) and its AUC, which reflects the proportion of false positives. We also introduce a mitigation rate (MR) to compare defense performance across heterogeneous environments, which is a normalized recovery ratio of a defended policy from the poisoned performance toward the clean one. These metrics provide a comprehensive evaluation of the method's performance and allow for comparison with baseline approaches. All evaluations were conducted with an RTX 5090 GPU.

\textbf{Baselines.}
We compare against representative backdoor defense baselines.
First, we implement direct retraining via PPO-based retraining \cite{schulman2017proximal} under the same environment and training budget for a fair comparison.
Second, we include Neural Cleanse (NC) \cite{wang2019neural}, STRIP \cite{gao2019strip}, and FeatureRE \cite{wang2022rethinking} as classic trigger-oriented defenses, adapted to RL policies by operating on state observations (as the model input) and flagging abnormal responses.
We further include RL-specific defenses, including BIRD\cite{chen2023bird}, SHINE\cite{yuan2024shine}, and PD\cite{bharti2022provable}.

\subsection{Experiment Results}


\begin{table*}[ht!]
\centering
\footnotesize
\caption{Performance of different defenses in clean and poisoned environments. We report the average episode return for SARL tasks and the average winning rate (\%) for MARL tasks over 1,000 rounds (higher is better). The best result in each column under each condition is bolded, and green shading marks competitive results close to the best.}
\label{tab:clean_poisoned_performance}
\vspace*{-2mm}
\begin{adjustbox}{max width=\textwidth} 
\begin{tabular}{ccccccccccc} 
\toprule
\multirow{2}{*}{\textbf{Environment}} & \multirow{2}{*}{\textbf{Methods}}
& \multicolumn{4}{c}{\textbf{SARL}}
& \multicolumn{3}{c}{\textbf{com-MARL}}
& \multicolumn{2}{c}{\textbf{coo-MARL}} \\
\cmidrule(lr){3-6}\cmidrule(lr){7-9}\cmidrule(lr){10-11}
&& \textbf{Breakout} & \textbf{SpaceInvaders} & \textbf{Seaquest} &  \textbf{Pong} &  \textbf{YSNP(\%)} &  \textbf{SH(\%)} &  \textbf{RTGA(\%)} &  \textbf{QMIX(\%)} &  \textbf{COMA(\%)} \\ \midrule
\multirow{10}{*}{\textbf{Clean}} & \textbf{Original} &445.3±125.2  & 722.2±248.2 & 1662.8±101.3 & 16.9±1.9 & 48.7±1.2 & 33.0±2.4 & 49.9±1.2 & 99.1±0.9 & 95.1±0.6 \\ 
& \textbf{Direct retraining} & 435.1±129.7 & \cellcolor{green!25}762.1±261.9 & - & 7.1±0.9 & 37.3±3.4 & 27.2±4.3 & 48.9±1.7 & \cellcolor{green!25}98.6±1.4 & \cellcolor{green!25}95.7±1.4 \\ 
& \textbf{NC} & 252.5±74.8 & 317.8±109.3 & - & 3.4±0.5 & - & - & - & \cellcolor{green!25}\textbf{98.8}±1.1 & \cellcolor{green!25}\textbf{96.1}±1.0\\ 
& \textbf{FeatureRE} & \cellcolor{green!25}450.1±131.4 & \cellcolor{green!25}741.8±255.1 & - & 7.3±1.0 & - & - & - & \cellcolor{green!25}\textbf{98.8}±1.2 & \cellcolor{green!25}95.7±1.0 \\ 
& \textbf{STRIP}& \cellcolor{green!25}476.8±138.3 & 707.3±243.1 & - & 15.2±2.0 & - & - & - & 87.1±1.1 & 79.3±1.7 \\ 
& \textbf{PD}& 212.2±80.9 & 281.8±154.1 & 901.5±122.3 & - & - & - & - & - & - \\ 
& \textbf{BIRD}& \cellcolor{green!25}478.2±156.0 & 626.3±260.9 & \cellcolor{green!25}\textbf{1684.1}±162.8 & - & 44.1±1.8 & 31.3±1.3 & 44.4±5.9 & 93.5±3.7 & 91.2±4.5 \\ 
& \textbf{SHINE} & \cellcolor{green!25}\textbf{505.5}±147.6 & \cellcolor{green!25}\textbf{880.1}±302.5 & - & \cellcolor{green!25}18.2±2.3 & \cellcolor{green!25}48.4±2.0 & \cellcolor{green!25}\textbf{37.2}±3.0 & \cellcolor{green!25}\textbf{50.8}±1.4 & \cellcolor{green!25}98.7±1.5 & \cellcolor{green!25}95.8±0.9 \\ 
& \textbf{BehaviorGuard} & \cellcolor{green!25}465.2±108.1 & \cellcolor{green!25}775.1±263.3 & \cellcolor{green!25}1672.8±86.5 & \cellcolor{green!25}\textbf{18.5}±1.6 & \cellcolor{green!25}\textbf{51.0}±3.1 & \cellcolor{green!25}33.3±4.3 & \cellcolor{green!25}49.4±0.5 & \cellcolor{green!25}98.5±1.2 & \cellcolor{green!25}95.3±0.8 \\ \midrule
\multirow{10}{*}{\textbf{Poisoned}} & \textbf{Original}& 33.1±81.7 & 114.1±236.1 & 141.2±46.4 & 2.1±3.7 & 18.3±2.6 & 15.9±1.0 & 18.8±2.3 & 21.9±0.2 & 35.7±0.7 \\ 
& \textbf{Direct retraining} & 309.0±156.7 & \cellcolor{green!25}752.0±280.9 & - & 3.0±3.4 & 27.7±4.1 & 27.8±3.1 & 37.4±1.2 & 82.6±1.1 & 82.3±3.2 \\ 
& \textbf{NC} & 23.2±69.0 & 396.7±183.6 & - & 0.1±4.2 & - &- & - & 25.9±0.4 & 21.6±0.5 \\ 
& \textbf{FeatureRE}  & 31.0±102.0 & 509.5±195.0 & - & 4.9±3.0 & - &- & - & 33.2±0.6 & 24.7±2.2 \\ 
& \textbf{STRIP} & \cellcolor{green!25}444.6±161.4 & \cellcolor{green!25}796.2±301.7 & - & 15.6±1.8 & - &- & - & 85.3±1.2 & 71.3±2.7 \\ 
& \textbf{PD} & 220.7±77.3 & 428.9±192.3 & 825.1±173.5 & - & - &- & - & - & - \\ 
& \textbf{BIRD} & 404.0±140.9 & 653.2±244.8 & \cellcolor{green!25}1630.8±162.9 & - & 43.6±3.9 &29.3±4.0 & 42.4±6.6 & \cellcolor{green!25}95.1±0.6 & \cellcolor{green!25}93.6±2.5 \\ 
& \textbf{SHINE} & \cellcolor{green!25}\textbf{570.9}±144.1 & \cellcolor{green!25}\textbf{951.5}±380.5 & - & \cellcolor{green!25}17.9±2.2 &\cellcolor{green!25}47.2±2.0 & \cellcolor{green!25}35.8±3.0 & \cellcolor{green!25}45.2±1.4  & \cellcolor{green!25}\textbf{98.6}±1.5 & 88.1±2.3 \\ 
& \textbf{BehaviorGuard} & \cellcolor{green!25}517.8±118.7 & \cellcolor{green!25}783.2±252.1 & \cellcolor{green!25}\textbf{1656.1}±106.7 & \cellcolor{green!25}\textbf{18.2}±1.5 &\cellcolor{green!25}\textbf{47.8}±1.1 & \cellcolor{green!25}\textbf{36.6}±2.6 & \cellcolor{green!25}\textbf{49.7}±1.1 & \cellcolor{green!25}96.9±1.7 & \cellcolor{green!25}\textbf{94.2}±1.1 \\ \bottomrule
\end{tabular}
\end{adjustbox}
\vspace*{-2mm}
\end{table*}

\subsubsection{Overall Defense Effectiveness}


Table~\ref{tab:clean_poisoned_performance} demonstrates the defense effectiveness of our BehaviorGuard and 7 baselines under the clean and poisoned environments. Among the 3 classic defenses, STRIP outperforms NC and FeatureRE, mainly because the latter two suffer from low-fidelity trigger reconstruction. Direct retraining can partially mitigate backdoors in SARL and coo-MARL, but fails in all com-MARL environments. BIRD and SHINE defend against backdoors via RL-specific trigger restoration and fine-tuning, but SHINE has stronger defense performance due to its high-fidelity trigger reconstruction. PD, as a retraining-free defense, performs poorly in both clean and poisoned environments due to its defective projection on all input states. BehaviorGuard does not introduce negative effects on the policy in the clean environment and achieves defense performance comparable to SHINE in the poison environment. While SHINE can accurately restore spatial triggers in SARL and coo-MARL, it struggles with complex sequential triggers used in com-MARL. In contrast, BehaviorGuard is trigger-agnostic and does not rely on trigger recognition, enabling consistent robustness across diverse RL settings. 

\subsubsection{Detection Performance}
\begin{figure}[!t]
\centering
\vspace{-5mm}
\begin{minipage}[c]{1\linewidth}
    \centering
    \subfloat[Single-agent]{\includegraphics[width=0.48\linewidth]{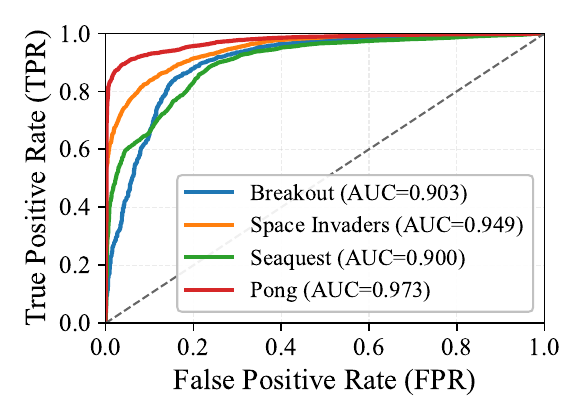}}
    \hspace{2mm}
    \subfloat[Multi-agent]{\includegraphics[width=0.48\linewidth]{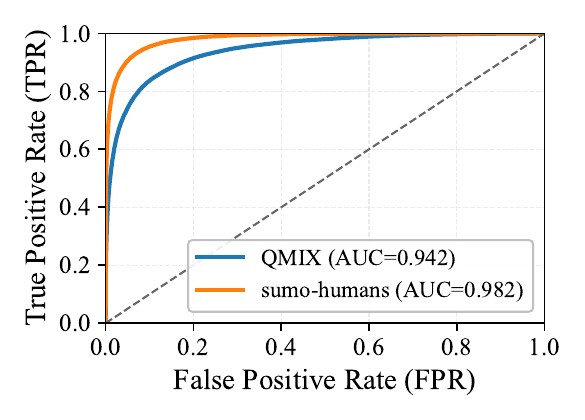}}
\end{minipage}
\vspace{-3mm}
\caption{ROC curves of BehaviorGuard's backdoor detection in (a) SARL Atari tasks and (b) multi-agent benchmarks.}
\label{fig:roc}
\end{figure}

We evaluated whether BehaviorGuard can reliably identify trigger activations. Fig.~\ref{fig:roc} demonstrates its detection quality for SARL and MARL via ROC curves and AUC values. Fig.~\ref{fig:roc}(a) shows that BehaviorGuard achieves consistently strong discrimination on SARL Atari tasks, with ROC curves concentrated toward the upper-left region and AUC values above 0.90 across games, indicating reliable separation between clean and backdoored policies under heterogeneous dynamics. Fig.~\ref{fig:roc}(b) further demonstrates that BehaviorGuard remains effective in multi-agent benchmarks, where interaction-induced non-stationarity and strategic coupling introduce substantial benign behavioral variability. Overall, these results indicate that our BDS provides a stable detection signal across single- and multi-agent systems.


\subsubsection{Robustness Across Advanced Attacks}
We further evaluated BehaviorGuard's defense effectiveness under four different advanced backdoor attacks: (i) \textbf{Normal attacks} (macro-averaged from Tab.~\ref{tab:clean_poisoned_performance}), (ii) \textbf{Reward-constrained attack} \cite{rathbun2025adversarial}, which enforces bounded backdoor reward modifications to implant backdoors; (iii) \textbf{Sequential attack in SARL} similar to \cite{yu2022temporal}, in which we use a patch to appear clockwise on the four corners of input images in 4 consecutive timesteps as a sequential trigger and perform backdoor actions after the 4 timesteps, and (iv) \textbf{Sequential attack in coo-MARL}, whose triggers are defined by agent-wise sequential actions with spatial and temporal dependences, similar to \cite{fang2025blast}. We use MR to compare defenses across heterogeneous environments and score scales. MR$=1$ indicates a clean-level recovery, and MR$>1$ allows slight over-recovery due to regularization effects.

Table~\ref{tab:mitigation_attack_types} compares robustness across attack types. Besides normal attacks, the other 3 advanced attacks suppress reward anomalies or embed triggers into sequential/interaction-dependent states, leading to low performance of trigger identification in BIRD and SHINE. BehaviorGuard achieves the most consistent robustness across attacks, matching strong baselines under normal attacks while notably outperforming existing methods under advanced attacks. 


\begin{table}[t]
\centering
\caption{Defense robustness across diverse backdoor attacks. We report MR (higher is better). Results are averaged over 1,000 evaluation episodes per environment and then macro-averaged across environments. S-Attack means sequential attack.}
\label{tab:mitigation_attack_types}
\vspace{-2mm}
\begin{adjustbox}{max width=\columnwidth}
\begin{tabular}{lcccc}
\toprule
\textbf{Method} &
\textbf{Normal} &
\textbf{Reward-Constrained} &
\textbf{S-Attack in SARL} &
\textbf{S-Attack in coo-MARL} \\
\midrule
BIRD  & 0.883$\pm$0.08 & 0.641$\pm$0.05 & 0.737$\pm$0.05 & 0.817$\pm$0.12 \\
SHINE & \textbf{1.074}$\pm$0.19 & 0.722$\pm$0.13 & 0.875$\pm$0.16 & 0.840$\pm$0.06 \\
Ours  & 1.054$\pm$0.09 & \textbf{1.090$\pm$0.06} & \textbf{0.981}$\pm$0.07 & \textbf{0.953}$\pm$0.05 \\
\bottomrule
\end{tabular}
\end{adjustbox}
\end{table}

\subsubsection{Time Cost of Backdoor Defense}
In Table \ref{tab:timecost}, we evaluated the defense time cost of SHINE, BIRD, PD, and BehaviorGuard in 4 single-agent environments. 
In each environment, following the standard workflow of these methods, we first ran a set of episodes (10000, 1000, 10, and 1000) with a maximum of 2,000 steps per episode to perform their offline operations and then ran 1,000 episodes to evaluate their online operations and defended policies. We measured the time cost of dominant phases in these methods. Existing methods require hours due to their expansive operations, including trigger restoration in SHINE, policy finetuning in SHINE and BIRD, and subspace sanitization and state projection in PD. BIRD's trigger restoration can converge early, thereby being faster than SHINE. PD and BehaviorGuard perform backdoor mitigation online, but the high computational cost of PD's online state projection limits its applicability for policy inference. In contrast, BehaviorGuard finishes in 35.03 minutes (18.81m for offline baseline calibration and 16.22m for online detection and mitigation), contributed by its lightweight drift scoring and action correction. In summary, BehaviorGuard significantly outperforms existing defenses in terms of both efficacy and time cost, and its online defense capability ensures practical applicability.


\begin{table}[t!]
\centering
\footnotesize
\caption{Time cost comparison. `None' denotes the basic inference time of a backdoor policy.}
\label{tab:timecost}
\vspace{-2mm}
\scalebox{0.94}{
\begin{tabular}{cP{0.45\columnwidth}cc}%
\hline
Method & Offline cost & Online cost & Total \\
\hline
None & --& 13.75m &13.75m\\\hline
\multirow{2}{*}{BIRD} & 12.0m (Trigger restoration) + 1.13h (Policy finetuning) & \multirow{2}{*}{13.76m} & \multirow{2}{*}{1.32h} \\
\multirow{2}{*}{SHINE} & 4.80h (Trigger restoration) + 1.95h (Policy finetuning) & \multirow{2}{*}{13.60m} & \multirow{2}{*}{6.75h} \\
PD & 0.93h (Subspace sanitization) & 1.29h & 2.22h \\
Ours & 18.81m (Baseline calibration) & 16.22m & \textbf{35.03m} \\
\hline
\end{tabular}}
\end{table}

\subsection{Ablation Studies}

\begin{figure}[!t]
\centering
\begin{minipage}[c]{1\linewidth}
    \centering
    \subfloat[Attack types]{\includegraphics[width=0.48\linewidth]{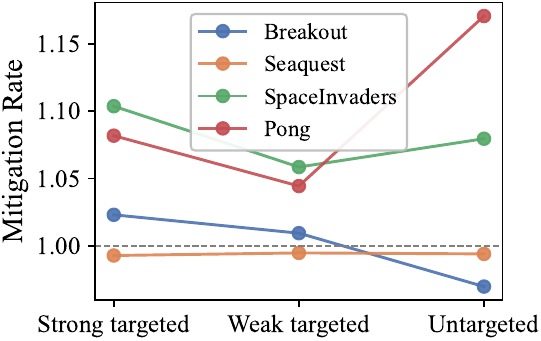}}
    \hspace{1mm}
    \subfloat[Guard probability $p$]{\includegraphics[width=0.48\linewidth]{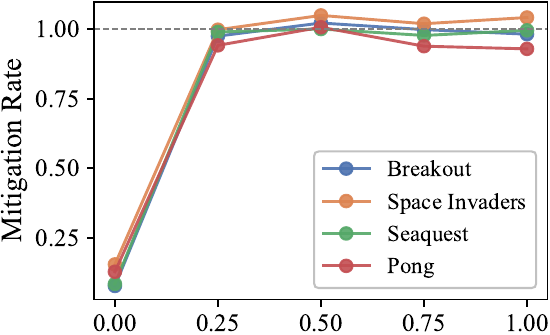}}
\end{minipage}

\vspace{-2mm}
\begin{minipage}[c]{1\linewidth}
    \centering
    \subfloat[Window length $L$]{\includegraphics[width=0.48\linewidth]{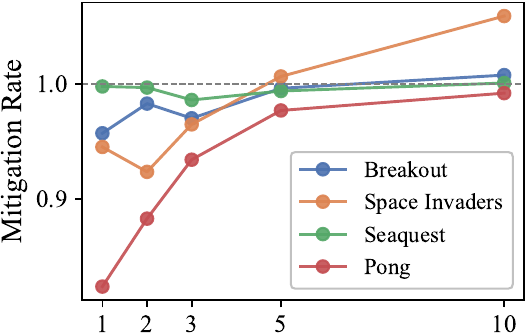}}
    \hspace{1mm}
    \subfloat[Trigger size]{\includegraphics[width=0.48\linewidth]{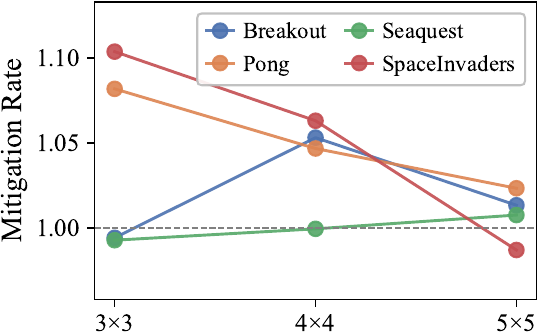}}
\end{minipage}
\vspace{-2mm}
\caption{Ablation studies of BehaviorGuard.
(a) Defense performance under different attack types, including strong targeted, weak targeted, and untargeted attacks.
(b) Effect of $p$, which controls the strength of probabilistic action correction once persistent drift is detected.
(c) Effect of $L$, determining how many consecutive drift detections are required to trigger mitigation.
(d) Defense robustness with respect to trigger size.
Results are averaged over multiple runs across Atari environments.}
\label{fig:ablation}
\end{figure}

We conducted ablations to assess the effectiveness and robustness of BehaviorGuard under different configurations. Results in Fig. \ref{fig:ablation} are averaged over multiple independent runs in Atari environments.

\textbf{Attack types.} Fig.~\ref{fig:ablation}(a) compares BehaviorGuard's MR under 3 TrojDRL attacks. Its performance under targeted attacks is more stable than under untargeted attacks, especially under strong targeted attacks. This indicates that BehaviorGuard is more reliable when the abnormal behavior manifests as sustained drift targeted by most backdoor attacks, rather than isolated, stochastic deviations.


\textbf{Guard probability $p$.} Fig.~\ref{fig:ablation}(b) studies the guard probability $p$ used in Eq.~\eqref{eq:tilde_a_t}, which controls the strength of probabilistic action correction after persistent drift is detected. Setting $p=0$ effectively disables mitigation and results in limited recovery, whereas moderate values (e.g., $p\in[0.25,0.5]$) provide a clear improvement across environments. Further increasing $p$ yields diminishing returns and may introduce unnecessary interventions. 

\textbf{Window length $L$.} Fig.~\ref{fig:ablation}(c) studies the impact of the window length $L$ used in Eq.~\eqref{eq:c_t}, \textit{i.e.}, the number of consecutive drift detections required to activate mitigation. Small $L$ values tend to trigger overly early corrections, while overly large $L$ delays intervention and reduces mitigation effectiveness. Intermediate $L$ achieves the best trade-off, supporting the design choice of modeling \emph{persistent} drift instead of reacting to instantaneous fluctuations. 

\textbf{Trigger size.} Fig.~\ref{fig:ablation}(d) evaluates robustness to trigger size. Mitigation performance remains stable across different trigger magnitudes without catastrophic degradation, indicating that the strategy does not rely on a specific trigger scale and generalizes across attack configurations. 


In summary, these results validate the necessity of combining temporal drift modeling with probabilistic action correction, and show that the key hyperparameters admit a reasonable operating range with consistent performance. More detailed sensitivity results are reported in Supplement B.

\subsection{Discussion and Limitations}

Our experimental results demonstrate behavioral drift as a trigger-agnostic signature for RL backdoor defense: it yields reliable separability across SARL and MARL and remains effective under interaction-induced non-stationary environments. BehaviorGuard's online guard achieves a favorable trade-off between utility and security, preserving clean performance while recovering poisoned performance without policy retraining or trigger reconstruction. The gains are especially clear for reward-constrained and sequential triggers where reward- or pattern-dependent defenses are weakened. BehaviorGuard still has a limitation. It is the reliance on benign baselines and a simple reference-action correction, which may require recalibration under distribution shift (e.g., new opponents or updated policies). Addressing this recalibration challenge and evaluating adaptive attackers that explicitly minimize drift are our future works.

\section{Conclusion}

We proposed a behavior-centric, online framework for backdoor detection and mitigation in DRL. Our key insight is that backdoored policies, despite differing trigger designs and objectives, tend to introduce persistent shifts in action distributions to ensure reliable activation, leaving detectable signatures in high-quantile regions and distribution tails even on trigger-free inputs. Based on this insight, we developed BehaviorGuard, which consists of a drift-based detector for backdoor activations without relying on reward anomalies, trigger priors, or trigger reconstruction, and a mitigator that suppresses backdoor actions at runtime and without requiring policy retraining. Extensive evaluations on SARL and MARL benchmarks demonstrate strong detection performance and a favorable utility-security trade-off: our defense substantially recovers poisoned performance while preserving clean-task performance, including under normal, reward-constrained, and sequential attacks. Future work will extend the drift constraint to continuous-action policies, improve calibration under distribution shift, and study adaptive attackers that explicitly optimize for drift evasion.

\section*{Acknowledgements}
This work was supported in part by the National Natural Science Foundation of China (62202387), the Guangdong Basic and Applied Basic Research Foundation (2025A1515011112), and the supporting funds for talents of Nanjing University of Aeronautics and Astronautics.

\bibliographystyle{named}
\bibliography{ref}
\appendix

\section{Defenses for MARL}
\label{asec:defenses}
\subsection{Behavior Drift Detection for MARL}

We extend the single-agent BDD to multi-agent settings by computing drift per agent and then aggregating it at the team level. In MARL, behavior is shaped by other agents, so the same policy can look different even without any trigger; this makes a fixed, single-agent-style criterion less reliable. To handle this, we add the Density Change Rate (DCR), which tracks how the drift-density evolves over time. Backdoor activation often causes an abrupt change in the drift distribution under interaction feedback, and DCR captures this effect even when marginal drift magnitudes are not strongly separated. We combine normalized drift magnitude, tail occupancy, and DCR into one detection score for both competitive and cooperative tasks.

\begin{algorithm}[t]
\caption{Multi-Agent Behavioral Drift Detection of BehaviorGuard}
\label{alg:ma-bdd}
\small
\begin{algorithmic}[1]
\Require
Action probabilities $\{p_{t,i,k}\}_{t=1}^{T_s}$ for agents $i\in\{1,\dots,n\}$; 
clean statistics $(\mu_{i,k},\sigma_{i,k})$;
KDE $(K,h)$; density threshold $T$; weights $(w_1,w_2,w_3)$; decision threshold $\tau$;\\
aggregation operator $\mathrm{Agg}(\cdot)$;
scenario flag $\textsc{Mode}\in\{\textsc{com},\textsc{coo}\}$; DCR thresholding parameters.
\Ensure Detection result for the evaluated policy set.
\State Initialize buffers $\mathcal{B}_i\gets[\,]$ for all agents $i$; 
$\mathcal{P}_i\gets[\,]$ for KDE densities.
\For{$t=1$ to $T_s$}
    \For{$i=1$ to $n$}
        \State $\text{BDS}_{t,i}\gets \frac{1}{2}\sum_{k=1}^{K_i}\left(\frac{p_{t,i,k}-\mu_{i,k}}{\sigma_{i,k}+\epsilon}\right)^2$
        \State Append $\text{BDS}_{t,i}$ to $\mathcal{B}_i$
    \EndFor
\EndFor

\For{$i=1$ to $n$}
    \For{$t=1$ to $T_s$}
        \State $\hat f_i(\mathcal{B}_i[t]) \gets \frac{1}{Nh}\sum_{j=1}^{N}K\!\left(\frac{\mathcal{B}_i[t]-\mathcal{B}_i[j]}{h}\right)$
        \State Append $\hat f_i(\mathcal{B}_i[t])$ to $\mathcal{P}_i$
    \EndFor
    \State $\text{tail}_i \gets \frac{1}{T_s}\sum_{t=1}^{T_s}\mathbf{1}\!\left[\hat f_i(\mathcal{B}_i[t])<T\right]$
    \State $\overline{\text{BDS}}_i \gets \frac{1}{T_s}\sum_{t=1}^{T_s}\mathcal{B}_i[t]$
\EndFor

\State \textbf{Team aggregation:} 
$\overline{\text{BDS}}_{\text{team}}\gets \mathrm{Agg}\big(\{\overline{\text{BDS}}_i\}_{i=1}^n\big)$, \;
$\text{tail}_{\text{team}}\gets \mathrm{Agg}\big(\{\text{tail}_i\}_{i=1}^n\big)$

\State Normalize with clean baselines:
$z_{\text{BDS}}\gets \frac{\overline{\text{BDS}}_{\text{team}}-\mu_{\text{BDS}}}{\max(\sigma_{\text{BDS}},\epsilon)}$, \;
$z_{\text{tail}}\gets \frac{\text{tail}_{\text{team}}-\mu_{\text{tail}}}{\max(\sigma_{\text{tail}},\epsilon)}$

\State \textbf{(Competitive only) Interaction-induced change:}
\For{$t=2$ to $T_s$}
    \State $\Delta_t \gets \mathrm{Agg}\big(\{\,|\mathcal{P}_i[t]-\mathcal{P}_i[t-1]|\,\}_{i=1}^n\big)$
\EndFor
\State $\overline{\Delta}\gets \frac{1}{T_s-1}\sum_{t=2}^{T_s}\Delta_t$
\State $z_{\text{dcr}}\gets \frac{\overline{\Delta}-\mu_{\text{dcr}}}{\max(\sigma_{\text{dcr}},\epsilon)}$

\State $\text{score}\gets w_1 z_{\text{BDS}} + w_2 z_{\text{tail}} + w_3 z_{\text{dcr}}$
\If{$\text{score}>\tau$}
    \State \Return ``Potential Backdoor Detected''
\Else
    \State \Return ``No Backdoor Detected''
\EndIf
\end{algorithmic}
\end{algorithm}

\subsection{Backdoor Mitigation for MARL}

Our mitigation follows the single-agent action guard but adapts the trigger condition and the intervention scope for MARL. Instead of reacting to a single abnormal step, we require persistent drift within a window to avoid false triggers caused by exploration and opponent-induced variance. The main difference from SARL is who gets corrected: in cooperative tasks, a single compromised agent can break coordination, so we allow a team-level gate; in competitive tasks, correcting the opponent is undesirable, so we use a victim-only gate. When the gate is on, we replace the raw action with a benign reference action with probability $p$, which provides a simple knob to trade off suppression strength and clean performance.

\begin{algorithm}[t]
\caption{Multi-Agent Detection-Guided Inference-Time Action Guard}
\label{alg:ma-guard}
\small
\begin{algorithmic}[1]
\Require
Policies $\{\pi_i\}_{i=1}^n$; per-agent drift scores $\{D_{t,i}\}$ (or BDS$_{t,i}$);
threshold $\tau_D$; window length $L$; guard probability $p$;\\
reference action sequences $\{\mathcal{A}_i^{\mathrm{ref}}\}$; aggregation operator $\mathrm{Agg}(\cdot)$;
scenario flag $\textsc{Mode}\in\{\textsc{com},\textsc{coo}\}$.
\Ensure Executed joint action $\tilde a_t=(\tilde a_t^1,\dots,\tilde a_t^n)$
\State Initialize buffers for drift indicators $z_{t,i}\gets 0$ and counters $c_{t,i}\gets 0$
\For{$t=1,2,\dots$}
    \State Observe observations $\{o_{t,i}\}_{i=1}^n$
    \For{$i=1$ to $n$}
        \State Raw action $a_{t,i}\gets \arg\max_a \pi_i(a\mid o_{t,i})$
        \State Reference action $a^{\mathrm{ref}}_{t,i}\gets \mathcal{A}^{\mathrm{ref}}_i[t \bmod T_i]$
        \State $z_{t,i}\gets \mathbb{I}(D_{t,i}>\tau_D)$
        \State Update $c_{t,i}\gets \sum_{j=\max(1,t-L+1)}^{t} z_{j,i}$
    \EndFor

    \If{$\textsc{Mode}=\textsc{coo}$}
        \State \textbf{Team gate:} $g_t \gets \mathbb{I}\!\left(\mathrm{Agg}(\{c_{t,i}\})\ge L\right)$
        \For{$i=1$ to $n$}
            \If{$g_t=1$ \textbf{and} $\mathrm{Unif}(0,1)<p$}
                \State $\tilde a_{t,i}\gets a^{\mathrm{ref}}_{t,i}$
            \Else
                \State $\tilde a_{t,i}\gets a_{t,i}$
            \EndIf
        \EndFor
    \Else
        \State \textbf{(Competitive) Guard victim subset $\mathcal{V}$ only.}
        \For{$i=1$ to $n$}
            \If{$i\in\mathcal{V}$ \textbf{and} $c_{t,i}\ge L$ \textbf{and} $\mathrm{Unif}(0,1)<p$}
                \State $\tilde a_{t,i}\gets a^{\mathrm{ref}}_{t,i}$
            \Else
                \State $\tilde a_{t,i}\gets a_{t,i}$
            \EndIf
        \EndFor
    \EndIf
    \State Execute $\tilde a_t$ and observe next state
\EndFor
\end{algorithmic}
\end{algorithm}

\section{Experiment Details}
\label{app:sensitivity}

\subsection{Ablation Study}
We examine whether the proposed Density Change Rate (DCR) is needed in competitive multi-agent settings. Experiments are conducted on three competitive tasks (YSNP/SH/RTGA) under the BACKDOORL attack. For each task, we evaluate on 1{,}000 clean episodes and 1{,}000 triggered episodes. We report detection AUC and TPR at a fixed operating point (FPR=1\%), and we also report the average win rate (WR) for clean execution (Clean), poisoned execution under triggers without defense (Poison), and triggered execution with the inference-time guard enabled (Mit.). Table~\ref{tab:dcr_ablation_com_marl} shows that the BDD-only variant performs notably worse, with lower AUC/TPR and weaker win-rate recovery. This is consistent with the fact that competitive MARL introduces interaction-driven non-stationarity, which increases the variability of single-agent drift statistics. DCR-only improves over BDD-only but remains behind the full method. Overall, BDD+DCR yields the most reliable detection at low FPR and the strongest recovery in win rate, indicating that DCR provides complementary interaction-dependent signals that are important for competitive MARL.

\subsection{BehaviorGuard under Diverse Attack Variants}
We further evaluate the robustness of \textbf{BehaviorGuard} under diverse backdoor variations in Atari environments, focusing on (i) different \emph{trigger sizes} and (ii) different \emph{attack types}. Unless otherwise specified, we follow the same evaluation protocol as in the main experiments and report the average episode score as mean$\pm$std over 1{,}000 episodes (higher is better). For each setting, we report \textbf{Clean} performance (trigger-free), \textbf{Poisoned} performance (triggered without defense), and \textbf{Mitigated} performance (triggered with BehaviorGuard enabled).

\paragraph{Robustness to Trigger Size.}
Table~\ref{tab:attack_size_env_scores} reports results under three commonly used patch sizes ($3{\times}3$, $4{\times}4$, and $5{\times}5$). As expected, backdoored policies suffer a severe performance collapse once the trigger is present, while clean performance remains stable across sizes. Importantly, BehaviorGuard consistently restores the poisoned performance across all four environments, and the recovered scores remain close to (or even exceed) the clean baseline in most cases. This indicates that our mitigation does not rely on a specific trigger magnitude and remains effective when the trigger becomes larger or visually more salient. We also observe minor fluctuations as size increases (e.g., in SpaceInvaders), which is expected since larger triggers can induce more frequent drift detections and thus stronger intervention; nevertheless, the overall mitigation effect remains stable.

\paragraph{Robustness to Attack Type.}
Table~\ref{tab:attack_type_env_scores} reports results under three TrojDRL-style attack types: \emph{Strong Targeted}, \emph{Weak Targeted}, and \emph{Untargeted}. The poisoned scores vary across these settings. Strong and weak targeted attacks typically enforce a clear failure pattern, leading to the most severe score drops, whereas untargeted attacks degrade behavior in a less structured manner and can leave higher poisoned scores in some games. In all cases, enabling BehaviorGuard improves the triggered performance. The recovery is most pronounced for strong/weak targeted attacks, consistent with our motivation that sustained, directional behavioral drift is easier to suppress at runtime. Although untargeted attacks can appear more stochastic and environment-dependent, BehaviorGuard still yields consistent gains while maintaining clean performance, indicating that the proposed guard is not tied to a single attack objective.

Overall, the results in Tables~\ref{tab:attack_size_env_scores} and~\ref{tab:attack_type_env_scores} show that BehaviorGuard maintains strong mitigation effectiveness across diverse backdoor variants, including variations in trigger size and attack objective, without requiring trigger priors or retraining.

\begin{table}[!t]
\centering
\caption{Mitigation Performance under Different Trigger Sizes across Atari Environments}
\label{tab:attack_size_env_scores}
\setlength{\tabcolsep}{6pt}
\renewcommand{\arraystretch}{1.12}
\begin{adjustbox}{max width=\columnwidth}
\begin{tabular}{@{}l l c c c@{}}
\toprule
\multirow{2}{*}{\textbf{Task}} &
\multirow{2}{*}{\textbf{Model}} & \multicolumn{3}{c}{\textbf{Trigger Size}} \\
\cmidrule(lr){3-5}
& & \textbf{3$\times$3} & \textbf{4$\times$4} & \textbf{5$\times$5} \\
\midrule

\multirow{3}{*}{Breakout}
& Clean     & $445.3\pm125.2$ & $445.3\pm125.2$ & $445.3\pm125.2$ \\
& Poisoned  & $33.1\pm81.7$   & $32.9\pm67.4$   & $32.4\pm79.6$ \\
& Mitigated & $\mathbf{517.8}\pm118.7$ & $\mathbf{534.2}\pm124.5$ & $\mathbf{522.4}\pm115.4$ \\
\midrule

\multirow{3}{*}{SpaceInvaders}
& Clean     & $722.2\pm248.2$ & $722.2\pm248.2$ & $722.2\pm248.2$ \\
& Poisoned  & $114.1\pm236.1$ & $105.4\pm142.7$ & $100.7\pm134.2$ \\
& Mitigated & $\mathbf{783.2}\pm252.1$ & $\mathbf{682.1}\pm234.8$ & $\mathbf{651.8}\pm185.7$ \\
\midrule

\multirow{3}{*}{Seaquest}
& Clean     & $1662.8\pm101.3$ & $1662.8\pm101.3$ & $1662.8\pm101.3$ \\
& Poisoned  & $141.2\pm46.4$   & $135.7\pm58.7$   & $168.5\pm49.8$ \\
& Mitigated & $\mathbf{1656.1}\pm106.7$ & $\mathbf{1689.4}\pm108.2$ & $\mathbf{1681.8}\pm128.4$ \\
\midrule

\multirow{3}{*}{Pong}
& Clean     & $16.9\pm1.9$ & $16.9\pm1.9$ & $16.9\pm1.9$ \\
& Poisoned  & $2.1\pm3.7$  & $5.2\pm1.2$  & $4.1\pm2.9$ \\
& Mitigated & $\mathbf{18.2}\pm1.5$ & $\mathbf{17.9}\pm1.4$ & $\mathbf{17.5}\pm2.3$ \\
\bottomrule
\end{tabular}
\end{adjustbox}
\end{table}

\begin{table}[!t]
\centering
\caption{Mitigation Performance under Different Attack Types across Atari Environments}
\label{tab:attack_type_env_scores}
\setlength{\tabcolsep}{6pt}
\renewcommand{\arraystretch}{1.12}
\begin{adjustbox}{max width=\columnwidth}
\begin{tabular}{@{}l l c c c@{}}
\toprule
\multirow{2}{*}{\textbf{Task}} &
\multirow{2}{*}{\textbf{Model}} & \multicolumn{3}{c}{\textbf{Attack Type}} \\
\cmidrule(lr){3-5}
& & \textbf{Strong Targeted} & \textbf{Weak Targeted} & \textbf{Untargeted} \\
\midrule

\multirow{3}{*}{Breakout}
& Clean     & $445.3\pm125.2$ & $445.3\pm125.2$ & $445.3\pm125.2$ \\
& Poisoned  & $33.1\pm81.7$   & $28.7\pm46.4$   & $212.5\pm64.7$ \\
& Mitigated & $\mathbf{517.8}\pm118.7$ & $\mathbf{511.4}\pm132.9$ & $\mathbf{495.4}\pm142.3$ \\
\midrule

\multirow{3}{*}{SpaceInvaders}
& Clean     & $722.2\pm248.2$ & $722.2\pm248.2$ & $722.2\pm248.2$ \\
& Poisoned  & $114.1\pm236.1$ & $98.4\pm135.8$ & $357.1\pm176.4$ \\
& Mitigated & $\mathbf{783.2}\pm252.1$ & $\mathbf{745.5}\pm250.1$ & $\mathbf{760.3}\pm257.4$ \\
\midrule

\multirow{3}{*}{Seaquest}
& Clean     & $1662.8\pm101.3$ & $1662.8\pm101.3$ & $1662.8\pm101.3$ \\
& Poisoned  & $141.2\pm46.4$   & $125.3\pm49.2$   & $716.6\pm63.1$ \\
& Mitigated & $\mathbf{1656.1}\pm106.7$ & $\mathbf{1670.0}\pm97.4$ & $\mathbf{1668.8}\pm83.2$ \\
\midrule

\multirow{3}{*}{Pong}
& Clean     & $16.9\pm1.9$ & $16.9\pm1.9$ & $16.9\pm1.9$ \\
& Poisoned  & $2.1\pm3.7$  & $1.5\pm0.9$  & $8.1\pm2.2$ \\
& Mitigated & $\mathbf{18.2}\pm1.5$ & $\mathbf{17.8}\pm2.6$ & $\mathbf{20.0}\pm0.8$ \\
\bottomrule
\end{tabular}
\end{adjustbox}
\end{table}

\begin{table}[t]
\centering
\caption{DCR Ablation in Competitive MARL}
\label{tab:dcr_ablation_com_marl}
\setlength{\tabcolsep}{4.5pt}
\renewcommand{\arraystretch}{1.10}
\begin{adjustbox}{max width=\columnwidth}
\begin{tabular}{@{}l l c c c@{}}
\toprule
\textbf{Task} & \textbf{Method} & \textbf{AUC} $\uparrow$ & \textbf{TPR@1\%FPR} $\uparrow$ & \textbf{WR(\%) Clean / Poison / Mit.} $\uparrow$ \\
\midrule
\multirow{3}{*}{\textbf{YSNP}}
& \textbf{Full (BDD+DCR)} & $0.98$ & $0.92$ & $48.7\;/\;18.3\;/\;\mathbf{50.2}$ \\
& w/o DCR (BDD-only)      & $0.84$ & $0.52$ & $48.7\;/\;18.3\;/\;34.0$ \\
& DCR-only                & $0.93$ & $0.78$ & $48.7\;/\;18.3\;/\;45.0$ \\
\midrule
\multirow{3}{*}{\textbf{SH}}
& \textbf{Full (BDD+DCR)} & $0.96$ & $0.88$ & $33.0\;/\;15.9\;/\;\mathbf{37.0}$ \\
& w/o DCR (BDD-only)      & $0.80$ & $0.45$ & $33.0\;/\;15.9\;/\;26.0$ \\
& DCR-only                & $0.90$ & $0.72$ & $33.0\;/\;15.9\;/\;32.0$ \\
\midrule
\multirow{3}{*}{\textbf{RTGA}}
& \textbf{Full (BDD+DCR)} & $0.97$ & $0.90$ & $49.9\;/\;18.8\;/\;\mathbf{50.8}$ \\
& w/o DCR (BDD-only)      & $0.82$ & $0.50$ & $49.9\;/\;18.8\;/\;36.0$ \\
& DCR-only                & $0.92$ & $0.75$ & $49.9\;/\;18.8\;/\;45.0$ \\
\bottomrule
\end{tabular}
\end{adjustbox}
\end{table}

\begin{table}[t]
\centering
\caption{Tail-Quantile $q$ Sensitivity under Backdoor Mitigation.}
\label{tab:t_density_sensitivity_qonly}
\begin{tabular}{lcccc}
\toprule
Game & $q$ & Clean & Poisoned & Mitigated \\
\midrule
\multirow{4}{*}{Breakout} 
 & 0.95  & 445.3 & 33.1 & 508.9 \\
 & 0.97  & 445.3 & 33.1 & 517.8 \\
 & 0.99  & 445.3 & 33.1 & 502.5 \\
 & 0.995 & 445.3 & 33.1 & 508.4 \\
\midrule
\multirow{4}{*}{SpaceInvaders} 
 & 0.95  & 722.2 & 114.1 & 783.2 \\
 & 0.97  & 722.2 & 114.1 & 763.3 \\
 & 0.99  & 722.2 & 114.1 & 741.2 \\
 & 0.995 & 722.2 & 114.1 & 749.8 \\
\midrule

 \multirow{4}{*}{Seaquest} 
 & 0.95  & 1662.8 & 141.2 & 1656.1 \\
 & 0.97  & 1662.8 & 141.2 & 1634.2 \\
 & 0.99  & 1662.8 & 141.2 & 1640.8 \\
 & 0.995 & 1662.8 & 141.2 & 1651.4 \\
\midrule
\multirow{4}{*}{Pong} 
 & 0.95  & 16.9 & 2.1 & 18.2 \\
 & 0.97  & 16.9 & 2.1 & 17.8 \\
 & 0.99  & 16.9 & 2.1 & 17.8 \\
 & 0.995 & 16.9 & 2.1 & 17.6 \\
\bottomrule
\end{tabular}
\end{table}

\end{document}